\documentclass[10pt,sigconf,letterpaper,nonacm,natbib=false]{acmart}
\usepackage{dirtytalk}
\usepackage{subcaption}
\AtBeginDocument{%
  }

\copyrightyear{2023}
\acmYear{2023}
\RequirePackage[
  datamodel=acmdatamodel,
  style=acmnumeric,
  ]{biblatex}

\begin{document}

% \title{Leveraging gradient-derived metrics for data selection and valuation in decentralised differentially private machine learning}
\title{Incentivising the federation: gradient-based metrics for data selection and valuation in private decentralised training}
% \author{Anonymous author}
% \author{Dmitrii Usynin}
% \author{Daniel Rueckert}
% \author{Georgios Kaissis}
\author{Dmitrii Usynin, Daniel Rueckert, Georgios Kaissis}
% \email{g.kaissis@tum.de}
\affiliation{%
  \institution{Technical University of Munich}
  \city{Munich}
  \country{Germany}
}
\affiliation{%
  \institution{Imperial College London}
  \city{London}
  \country{United Kingdom}
}

\renewcommand{\shortauthors}{Usynin et al.}

\begin{abstract}
Obtaining high-quality data for collaborative training of machine learning models can be a challenging task due to A) regulatory concerns and B) a lack of data owner incentives to participate. The first issue can be addressed through the combination of distributed machine learning techniques (e.g. federated learning) and privacy enhancing technologies (PET), such as the differentially private (DP) model training. The second challenge can be addressed by rewarding the participants for giving access to data which is beneficial to the training model, which is of particular importance in federated settings, where the data is unevenly distributed. However, DP noise can adversely affect the underrepresented and the atypical (yet often informative) data samples, making it difficult to assess their usefulness. In this work, we investigate how to leverage gradient information to permit the participants of private training settings to select the data most beneficial for the jointly trained model. We assess two such methods, namely variance of gradients (VoG) and the privacy loss-input susceptibility score (PLIS). We show that these techniques can provide the federated clients with tools for principled data selection even in stricter privacy settings.
\end{abstract}

% <concept_id>10002978.10003029.10011150</concept_id>
% <concept_desc>Security and privacy~Privacy protections</concept_desc>
% <concept_id>10002978.10002991.10002995</concept_id>
% <concept_desc>Security and privacy~Privacy-preserving protocols</concept_desc>

% \ccsdesc[500]{Security and privacy~Privacy protections}
% \ccsdesc[500]{Security and privacy~Privacy-preserving protocols}



\maketitle

\section{Introduction}
To train machine learning (ML) models which are effective at solving various tasks, the model owner(s) need descriptive, diverse and well-curated training data. However, in many sensitive contexts such data (e.g. patients with rare conditions) can be very challenging to get \cite{ahmad2020fairness, mhasawade2021machine, glocker2023algorithmic}. Currently, various strategies can be leveraged to obtain this data defined as collaborative machine learning (CML), but these techniques still suffer from two major problems: data privacy requirements and a lack of incentive to participate. 

Firstly, a number of data protection and governance regulations (such as GDPR) stipulate that the collection and usage of sensitive data should be minimised.
One solution to this problem was proposed in \cite{konevcny2016federated}, namely federated learning (FL). However, FL on its own fails to account for the adversarial actors in collaborative settings, making it vulnerable to privacy attacks \cite{geiping2020inverting, usynin2022beyond}. One method commonly used to alleviate these issues is differentially private (DP) training \cite{dpbook}, often in the form of differentially private stochastic gradient descent (DP-SGD) \cite{abadi2016deep}, which is capable of providing objective and quantifiable guarantees of privacy for each client.

\begin{figure}[!h]
\centering
\includegraphics[width=0.75\linewidth]{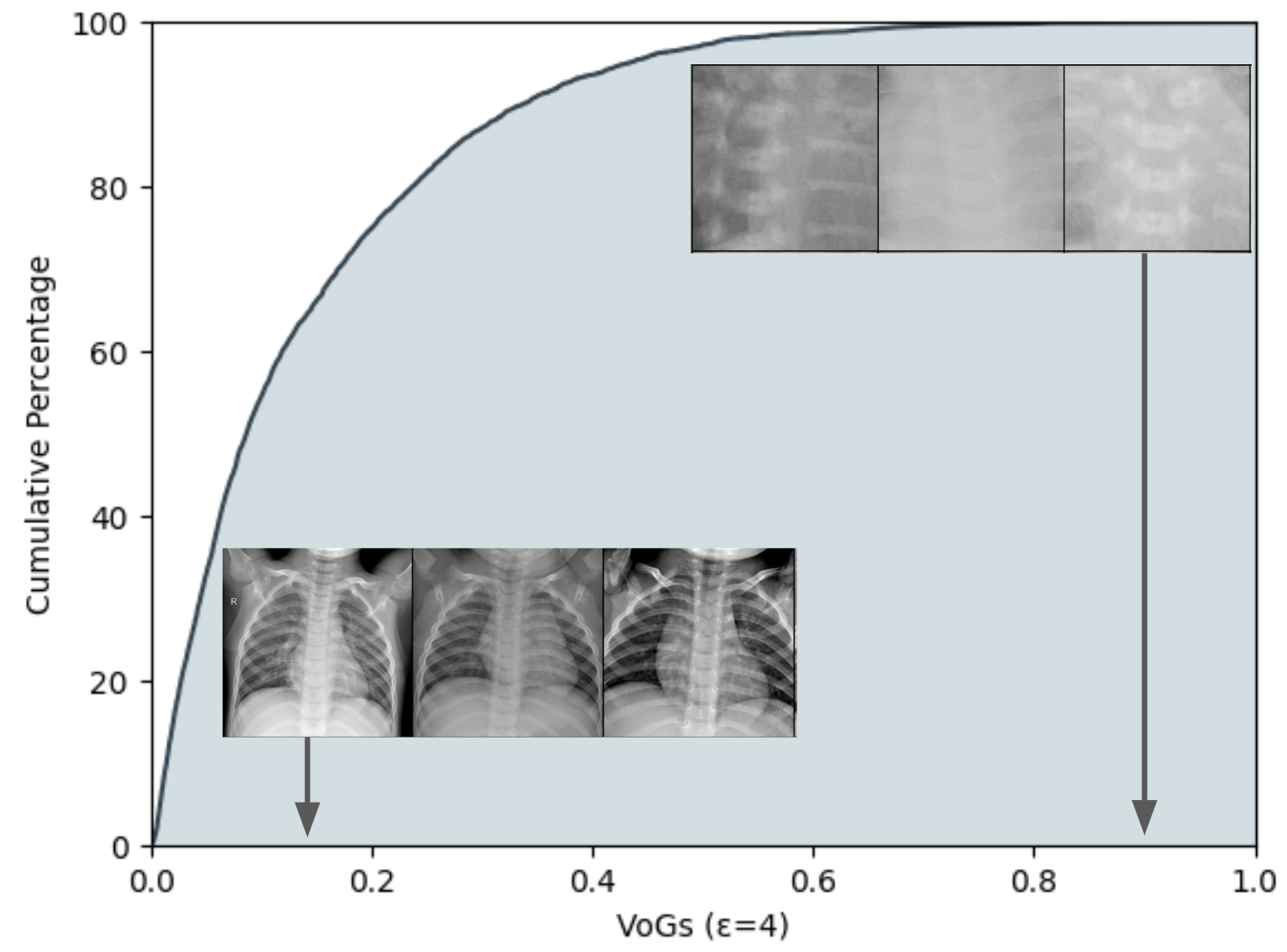}
\caption{Distribution of normalised VoG scores for ResNet-$18$ (PPPD, $\varepsilon$=$4$). Higher values indicate atypical samples.}
\label{fig:vog_example}
\end{figure}

Secondly, in order to train such models, data of high utility needs to be selected.
This is often data which contains more features of interest or is scarce.
However, it is often challenging to obtain access to such data samples.
The problem is twofold: Currently there is no general agreement on how to establish how much individual data samples are worth and the model owners do not have the incentive to pay for them unless their model is likely to improve from it.
To alleviate these issues, the relative usefulness of the data can be quantified and used to reward the participants.
Many of these methods have been adapted to FL, permitting analysis of the values associated with individual data points without needing to inspect the underlying data, such as \cite{xu2021gradient} or \cite{agarwal2022estimating}, which can help model owners to guide the training process and reward the participants appropriately.

While both of these problems can be reasonably alleviated individually, addressing both of them simultaneously can be challenging.
DP often affects the utility of the resulting model \cite{bagdasaryan2019differential}, rendering many training analysis techniques infeasible. 
And as a result, it can be challenging to objectively assess the contributions of each individual participant.
Therefore, many commonly employed data attribution techniques, such as per-sample loss values or gradient norms \cite{amiri2021convergence, lai2021oort} often fail to objectively assess the individual contributions in DP settings. 
Additionally, DP training was shown to result in biased models, which perform worse on underrepresented subgroups \cite{farrand2020neither}, further reducing the diversity of the shared contributions.
Methods such as data valuation using Shapley Values \cite{jia2019towards, sim2020collaborative} can often be rendered ineffective due to gradient clipping, reducing the contributions for diverse and out-of-distribution (OOD) samples \cite{jia2021scalability}. Moreover, as computational performance can be critical in many collaborative settings, DP training itself can add a significant performance overhead.
As a result, methods such as leave-one-out (LOO) \cite{jia2021scalability}, which requires retraining the model multiple times or calculation of Shapley values can be too costly to employ.
% , further inflating the costs of such collaborations.

In this work we identify a number of methods which can be used to select \say{useful} training samples for both normal and DP training.
While there exists a number of definitions of sample usefulness, in this work we refer to samples, which, upon exclusion from the dataset, lead to poorer model generalisation (similarly to the definition of \cite{baldock2021deep}).
One method to achieve this is the Variance of Gradients (VoG) \cite{agarwal2022estimating}, which allows the participants to evaluate the difficulty of individual data samples, and select data points that are more challenging for the model to learn from (we demonstrate a range of such samples in Figure \ref{fig:vog_example}) and can, thus, benefit generalisation.
In this work, the notions of image \textit{difficulty} and \textit{atypicality} are used identically to \cite{agarwal2022estimating}.
To improve training data diversity, we employ the the privacy loss-input susceptibility (PLIS) \cite{mueller2022input} score. 
Typically, this metric is used to identify more privacy-sensitive (we also refer to them as \textit{revealing} in this work) attributes and data points, however, it has also previously been used to detect atypical/OOD samples in sensitive datasets and is explicitly formulated for DP training.
Both of these metrics can also be conveniently realised as easily comparable scalar values.
This, in turn, simplifies the identification of more valuable samples and a reward allocation process. 
Through extensive experimental results we show that these metrics can be used to augment the existing FL settings with both privacy and incentivisation in mind.

Overall, the contributions of this work can be summarised as follows:
\begin{itemize}
    \item We investigate how gradient-based metrics can be used in collaborative DP machine learning to identify samples of higher utility and compare them to commonly used sample utility metrics
    \item We provide initial evidence that gradient-based metrics can be used to effectively identify samples of interest in DP and non-DP settings across a variety of model architectures, datasets and privacy regimes
    \item As these metrics are themselves privacy-sensitive, we propose a DP version of these metrics that can be shared with other participants. This allows the federation to identify sites with non-standard datasets and individuals, who can be reimbursed for access to their data
\end{itemize}

\section{Background and related work}
\label{sec:related}
\subsection{Differential privacy}
There exists a large number of works on PETs deployed in low-trust collaborative environments \cite{usynin2021adversarial}.
% In this work we concentrate on differential privacy specifically. 
In this work we are specifically interested in a method which would allow the federation to retain their \textit{output} privacy \cite{usynin2021distributed}, which entails protection of their sensitive data (and its derivatives) from adversaries who are part of the training consortium or are able to query the model that has already been trained. As the the definition of \textit{input} privacy has already been satisfied through the use of FL, we, thus study the concept of differential privacy used to ascertain the privacy of the information stored in the shared model updates.
\begin{definition}[$(\varepsilon, \delta)$-DP, \cite{dpbook}]
We say that the randomised mechanism $\mathcal{M}$ preserves $(\varepsilon, \delta)$-DP if, for all pairs of adjacent databases $D$ and $D'$ and all subsets $\mathcal{S}$ of $\mathcal{M}$'s range:
\begin{equation}
    \mathbb{P}(\mathcal{M}(D) \in \mathcal{S}) \leq e^{\varepsilon} \, \mathbb{P}(\mathcal{M}(D') \in \mathcal{S}) + \delta.
\end{equation}
\end{definition}

Most properties of $(\varepsilon, \delta)$-DP are described in detail in \cite{dpbook}.
In simple terms, the output of a DP algorithm is approximately invariant to an inclusion or an exclusion of an individual.

\subsection{Data valuation}
In this work we are interested in determining the importance or value of individual data points.
Most prior works in this area rely on the aforementioned Shapley values or LOO retraining in order to determine the usefulness of the shared data \cite{jia2021scalability, ghorbani2019data, shobeiri2022shapley}.
However, as our setting employs DP, some of the data, particularly the outliers, can be adversely affected by gradient noise when approximating the importance of individual data points \cite{chen2020understanding, farrand2020neither}. 
Additionally, these calculations are very computationally expensive, resulting in large overhead during training, which can render these methods infeasible in many settings. 

\section{Methods}
In this work we employ the variance of gradients method described in \cite{agarwal2022estimating}. Formally, pixelwise $\text{VoG}_{p}$ is defined as follows:
\begin{equation}
    \text{VoG}_p = \sqrt{\frac{1}{K}}\sum^{K}_{t=1}(S_t - \mu)^2,
\end{equation}
where $\mu = \frac{1}{K} \sum^{K}_{t=1}S_t$, $K$ denotes time steps, $N$ denotes input pixels and $S_t$ is the gradient of the loss with respect to the input pixels. The scalar VoG is calculated from the averaged pixel-wise variance of gradients over all time steps as $\frac{1}{N}\sum^{N}_{t=1}(\text{VoG}_p)$. Similarly to \cite{agarwal2022estimating}, we perform per-class normalisation. However, unlike \cite{agarwal2022estimating}, we normalise the values to lie in the region of $[0,1]$ for easier interpretability.

Additionally, we also employ PLIS \cite{mueller2022input}, which is defined as follows:
\begin{align}
    \mathbf{PLIS}(S_i) &\coloneqq \boldsymbol{J}_{x_i} (\mathbf{PL}(S_i))  \nonumber \\
                &= \boldsymbol{J}_{x_i} \left(\frac{\Vert \nabla_{\boldsymbol{\theta}} \ell(x_i | y_i, \boldsymbol{\theta}) \Vert_2^2}{\sigma^2}\right).
\label{eq:PLIS}
\end{align}

Where $\mathbf{PL}(S_i)$ is:   

\begin{equation}
    \mathbf{PL}(S_i) \propto \frac{\Vert \nabla_{\boldsymbol{\theta}} \ell(x_i | y_i, \boldsymbol{\theta}) \Vert_2^2}{\sigma^2}.
\end{equation}

Here $J_{x_{i}}$ is a Jacobian with respect to the input attribute (pixel) $i$; $\ell(f(x, \theta),y)$ is the loss function of a neural network $f(x, \theta)$, where $(x_i, y_i)$ is a input/label pair belonging to subject $S_i$, and $\boldsymbol{\theta}$ is a vector of learnable parameters

In this work we report the spectral norm ($L_2$-norm) of the PLIS matrix, which allows the individuals to utilise a single scalar value to approximate their privacy-loss with the respect to the input features. We normalise the class-wise PLIS scores to lie in the region of $[0,1]$ for easier interpretability.

% By employing VoG and PLIS, we can identify the samples which are A) more diverse and B) more difficult for the model to learn from, potentially making them more desirable in a collaborative learning setting. PLIS can be employed in conjunction with VoG to detect the OOD training samples and further leverage the same gradient we have previously computed for each individual. Both of these values are considered to be private data and should not be shared with the rest of the federation in a non-private form. We discuss this in detail in Section \ref{sec:dp_vog}.

We provide the descriptions of our training settings, including models and datasets used, hardware and privacy regimes in Appendix \ref{sec:settings}.

\section{Results and discussion}
\begin{figure}[ht!]
\centering
      \includegraphics[width=0.45\linewidth]{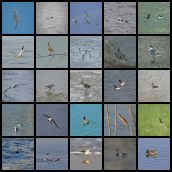}
      \hspace{5mm}%
      \includegraphics[width=0.45\linewidth]{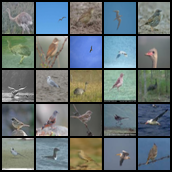}
\caption{Comparison of images with largest VoGs for WideResNet-$50$ ($\varepsilon$=$4$) and ResNet-$101$ ($\varepsilon$=$4$) respectively (CIFAR-$10$, \textit{bird} class). The trend for low contrast and more defined features is maintained across different model architectures (SSIM of $0.552$ and BD of $0.698$).}
\label{fig:cifar_models}
\end{figure}

We find that, in general, the selected images between the DP and non-DP settings are very similar (Figure \ref{fig:cifar_models}).
In both settings, the samples selected are atypical and represent the data that can be considered difficult for a model to learn from.
We now discuss how VoG and PLIS values can be affected in different learning settings and how they compare against other commonly used metrics.

% We discuss the relationship between VoG and PLIS in detail in Section \ref{sec:plis_comp}.

Our main findings can be summarised as follows:
\begin{itemize}
    \item Unlike most of the commonly used techniques, VoG-based sample selection is consistent across many model architectures, datasets and privacy regimes;
    \item Larger models are, in general, more difficult to analyse, as the sample selection is not consistent for larger datasets;
    \item While both PLIS and VoG scores can be utilised to identify samples of interest, they account for different information content and can be used to prioritise different sample types.
\end{itemize}

We quantify the selection \say{consistency}, using the structural similarity index (SSIM) \cite{wang2004image} and the Bhattacharyya distance (BD) \cite{kailath1967divergence} between the pixel distributions of the images. Higher SSIM and lower BD distance indicate similar images.
Alternative metrics can be employed instead (e.g. the Fretchet Inception distance \cite{heusel2017gans}).

\begin{figure}[ht!]
\centering
      % \includegraphics[width=0.3\linewidth]{figures/w50_vanilla_cl2.png}
    % \hspace{5mm}%
      \includegraphics[width=0.8\linewidth]{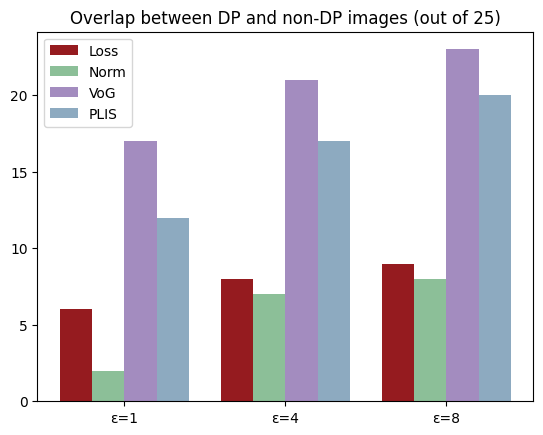}
\caption{Comparison between different selection methods under normal and DP training (ResNet-$18$, CIFAR-$10$). Higher is better.}
\label{fig:graphs}
\end{figure}

\subsection{Comparing against per-sample losses and gradient norms}
\begin{figure}[!h]
\centering
\begin{subfigure}[]{0.5\textwidth}
    \includegraphics[width=0.45\linewidth]{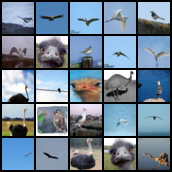}
      % \hspace{5mm}%
      \includegraphics[width=0.45\linewidth]{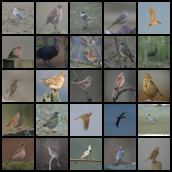}
\end{subfigure}
    \begin{subfigure}[]{0.5\textwidth}
        \includegraphics[width=0.75\linewidth]{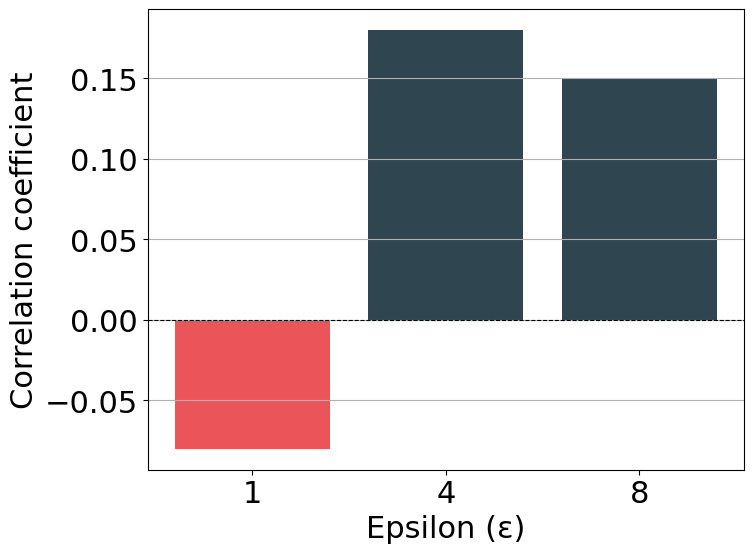}
    \end{subfigure}
      
\caption{Comparison of images with largest gradient norms for DP and non-DP models respectively (ResNet-$18$, CIFAR-$10$, \textit{bird} class, $\varepsilon=4$). There is little conceptual similarity between the chosen images (low correlation coefficients at different $\varepsilon$ values, the SSIM of $0.355$ and the BD of $0.914$).}
\label{fig:norms}
\end{figure}

We firstly compare the metrics we selected against the commonly used sample selection strategies in FL: per-sample loss and $L_2$-gradient norm values.
Both of these are often used as proxies for estimation of sample difficulty in FL \cite{amiri2021convergence, lai2021oort, li2021sample, xue2021toward}. 
However, due to the addition of noise and aggressive gradient clipping, the selected images are not consistent in DP settings (exemplified in Figure \ref{fig:graphs}). 
We notice that there is a large amount of variation between the images in the private and non-private settings (Figure \ref{fig:norms}), and in some cases there is almost no overlap between the two.
The Pearson correlation coefficients of the top $1000$ private ($\varepsilon=4$) and non-private scores were $0.09$ and $0.18$ (loss values and gradient norms respectively).
In comparison (discussed below), both the VoG and PLIS scores are consistent across most models, datasets and privacy levels. 
Therefore, while loss- and norm-based metrics can identify similar images compared to VoG and PLIS scores in non-private settings, they are not suitable for DP training.

\subsection{Increasing the model size}
\begin{figure}[!h]
\centering
      \includegraphics[width=0.37\linewidth]{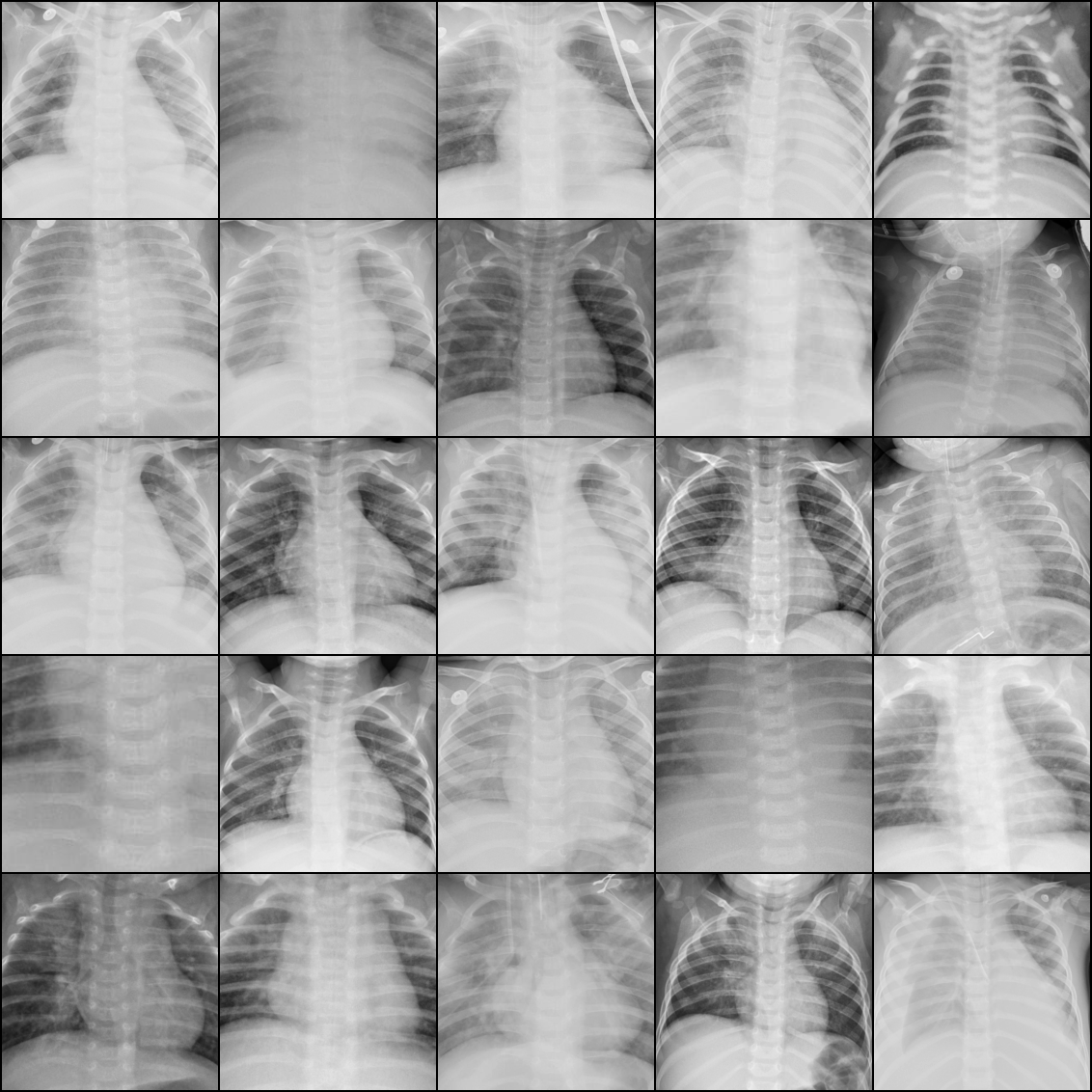}
    \hspace{5mm}%
      \includegraphics[width=0.37\linewidth]{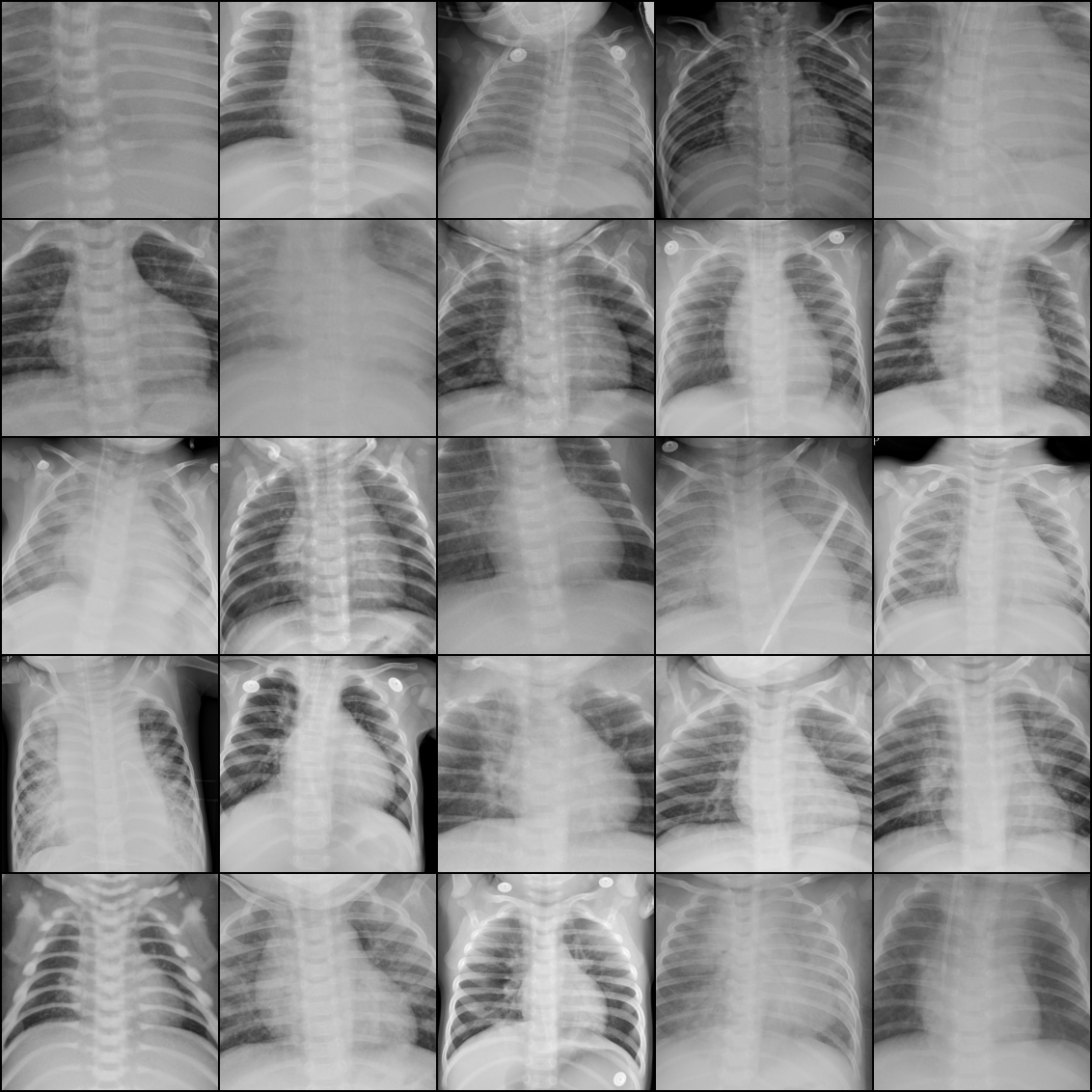}
\caption{Comparison of images with largest VoGs for ResNet-$18$ and ResNeXt-$101$ respectively (non-private models PPPD). Here there is some variation in the images (SSIM of $0.521$ and BD of $0.320$) based on the size of the model even in non-private settings.}
\label{fig:pppd_size_vanilla}
\end{figure}

We observe that, in general, regardless of the number of parameters (and of the architecture) the VoG scores can be used to successfully identify the atypical samples in lower-dimensional datasets (Figure \ref{fig:cifar_models}).
However, we also notice that larger models tend to be more challenging to analyse.
We show exemplary results in Figure \ref{fig:pppd_size_vanilla} and note that there is a significant difference between the images of interest for ResNet-$18$ and ResNeXt-$101$ for both the DP and non-DP models.
We notice that as the model size grows, for PPPD the usefulness of each individual sample becomes more difficult to define and the selected images feature both highly-detailed X-rays as well as blurry images with an undersized field of view.
This could make the adoption of VoG-based methods trickier for settings that employ models with a larger number of parameters.
% So far, we have only analysed ResNet-based architectures, while the investigation of the interplay between different architectures and the number of model parameters is part of our ongoing work. 
To summarise: We tend to see very consistent sample selection for smaller images, but for larger images, the attribution is more challenging to interpret.

\subsection{Effects of changing the $\varepsilon$-value}
\begin{figure}[h!]
\centering
    %   \includegraphics[width=0.45\linewidth]{figures/r18_cif_e1_class0.png}
    % \hspace{5mm}%
      \includegraphics[width=0.5\linewidth]{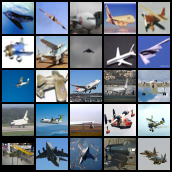}
\caption{Images with largest VoGs for $\varepsilon$ of $1$ and $8$ (ResNet-$18$, CIFAR-$10$, \textit{airplane} class). Selected samples are identical with an SSIM of $1.0$.}
\label{fig:cifar_epsilon}
\end{figure}

% Overall, while there could be small variations in the ordering of the selected images at different levels of $\varepsilon$, 
We discovered that the atypical images identified with VoG scores are consistent across all privacy levels for lower-dimensional datasets, showing that participants can effectively employ VoGs for sample selection with strong privacy guarantees.
% Thus, the federation can benefit from this sample selection method even under the strictest privacy regimes.
We show exemplary results in Figure \ref{fig:cifar_epsilon}, where the images selected across all privacy levels are almost identical (with SSIM being close to $1.0$).
However, for PPPD, we observe that there is more variation in image selection.
We note that the trend of atypical images being considered to be more difficult remains the same, but the features which make these images stand out differ.
We show these results in Figure \ref{fig:pppd_epsilon} and hypothesize that there could be two factors at play here.
Firstly, the size and the complexity of the dataset, which we discuss in detail below.
Secondly, the accuracy of the model: For smaller images, the accuracy of the trained model does not seem to have an impact in private settings, but for PPPD (e.g. the accuracy difference between $\varepsilon$ of $1$ and $8$ on ResNet-$18$ is $21.1\%$) the results vary severely.
One interesting discovery is that for a lower value of $\varepsilon$, the selected images seem to be more atypical, showing that models of lower accuracies could be used to identify images, which a human observer would also consider to be more unusual.
We hypothesise that this might be due to the fact that a more private model might only concentrate on the \say{core} image features.

A similar observation can be made about the relationship between the value of $\varepsilon$ and the samples of interest detected using the PLIS score.
In moderate-to-low noise settings ($\varepsilon$ of $4$ and $8$ respectively), most privacy-sensitive samples remain fairly consistent (with SSIMs in a range of $0.8$).
However, for $\varepsilon$ of $1$, we observe a very different set of selected images. 
This could have a similar implication of the model only concentrating on the most revealing features of the input image, or be due to an over-proportional decrease in memorisation ability at low privacy values, as discussed in \cite{feldman2020does}.
As part of future work, these results could help us find the answer to a more fundamental question: \say{If a model is restricted to only learning from a small subset of features, which features would it learn first?} 

\subsection{Changing the dataset}
\begin{figure}[h!]
\centering
      \includegraphics[width=0.45\linewidth]{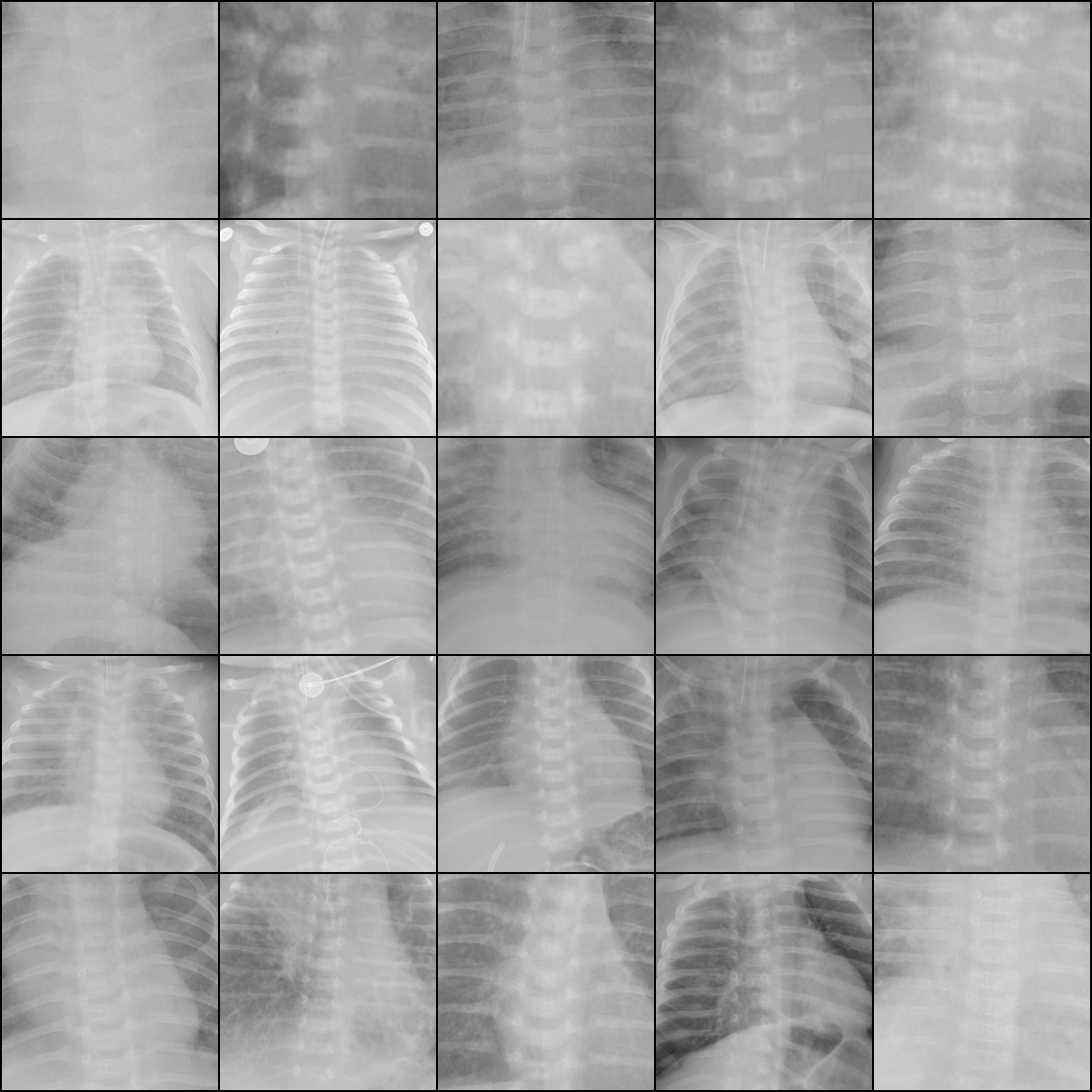}
    \hspace{5mm}%
      \includegraphics[width=0.45\linewidth]{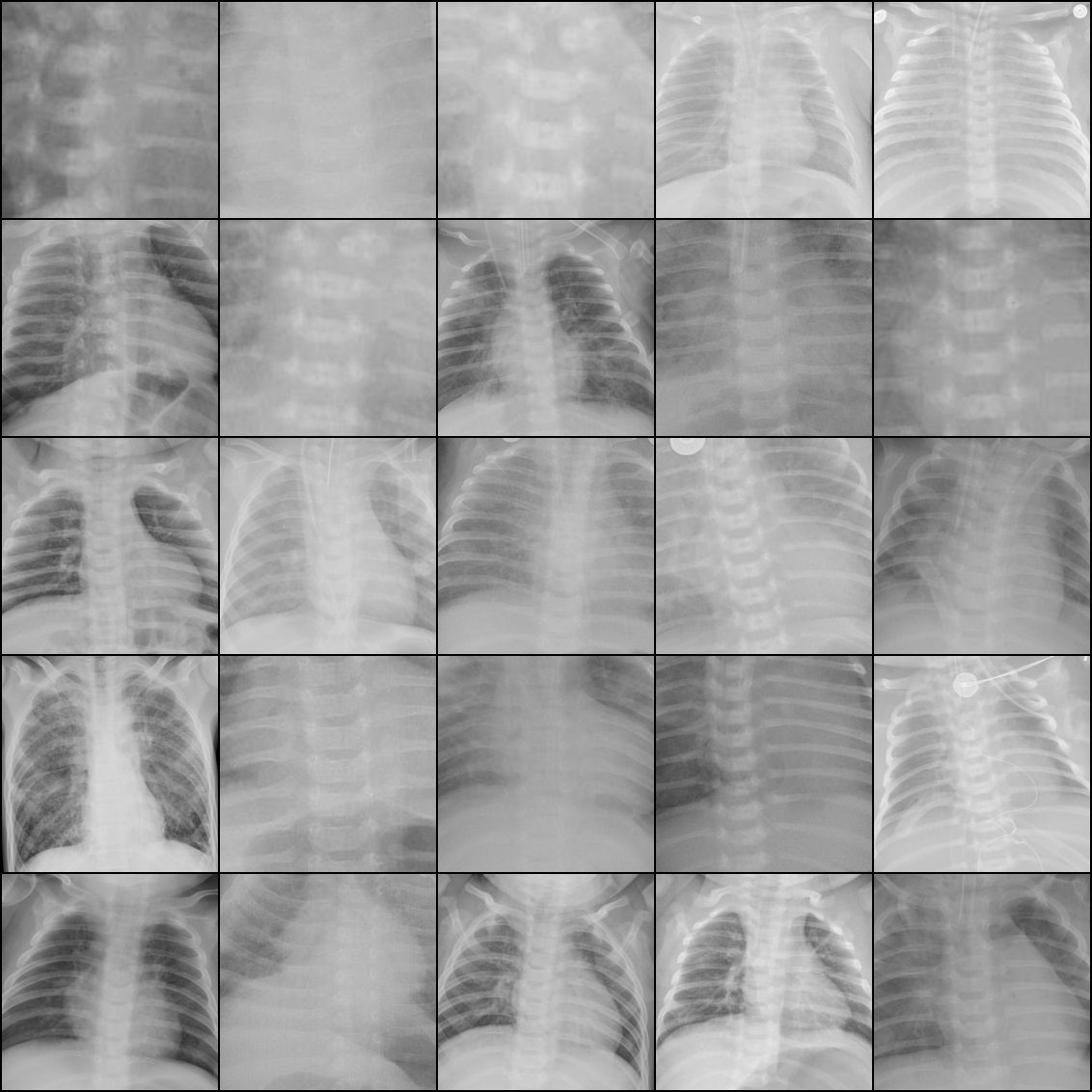}
\caption{Comparison of images with largest VoGs for $\varepsilon$ of $1$ and $8$ respectively (ResNet-$18$, PPPD). The overall selection criteria (low contrast and unclear anatomical details) remains consistent (SSIM of $0.722$ and BD of $0.074$).}
\label{fig:pppd_epsilon}
\end{figure}

\begin{figure}[!h]
\centering
\includegraphics[width=0.4\linewidth]{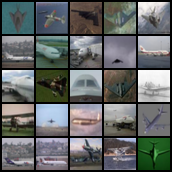}
\hspace{5mm}%
\includegraphics[width=0.4\linewidth]{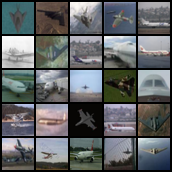}
\caption{Comparison of images with largest VoGs for $\varepsilon$ value of $4$ and $8$ respectively (ResNet-$9$, CIFAR-$10$, \textit{airplane} class). $21/25$ images are identical.}
\label{fig:eps_id}
\end{figure}

Both the small and the large image settings showed that VoGs could be successfully leveraged to identify atypical samples across all values of $\varepsilon$ regardless of the dataset. 
We note that PPPD showed more variation of samples of interest across different model architectures and $\varepsilon$ values.
We hypothesise that this can be the case either due to A) PPPD much smaller (only $5\;400$ samples), or B) only containing a single input channel, resulting in a more profound role of the contrast of the image.
Thus, we see that regardless of the quality of the individual samples, the higher values identify images which show lower contrast and level of detail.
We show exemplary results for the largest and the smallest VoGs for PPPD in Figure \ref{fig:pppd_epsilon}.
Finally we note, that as the dataset size grows (e.g comparing CIFAR-$10$ and CINIC-$10$), the selection consistency grows as well with $21$ and $24$ overlapping samples (out of $25$) for $\varepsilon$=$4$ respectively. 
We show exemplary results in Figure \ref{fig:eps_id}. 

\subsection{Relationship between PLIS and VoGs}
\label{sec:plis_comp}
\begin{figure}[h!]
\centering
      \includegraphics[width=0.45\linewidth]{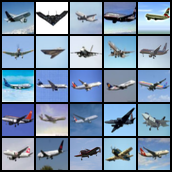}
    \hspace{5mm}%
      \includegraphics[width=0.45\linewidth]{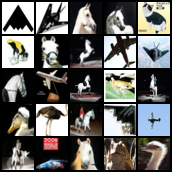}
\caption{Images with the smallest VoG and PLIS scores respectively (ResNet-$18$, $\varepsilon=1$, CIFAR-$10$). The selection of images is severely different based on which metric is used (SSIM of $0.215$ and BD of $0.886$).}
\label{fig:vog_plis_dp}
\end{figure}

\begin{figure}[!h]
\centering
      \includegraphics[width=0.45\linewidth]{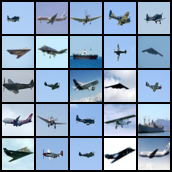}
    \hspace{5mm}%
      \includegraphics[width=0.45\linewidth]{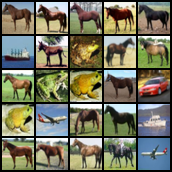}
\caption{Comparison of images with smallest PLIS for private ($\varepsilon=4$) and non-private settings respectively (ResNet-$18$, CIFAR-$10$, all classes). The dominant classes are different.}
\label{fig:plis_priv_nonpriv}
\end{figure}

When comparing the PLIS and VoG scores, we notice that there no immediate correlation between them ($0.57$ and $0.55$ values of Pearson correlation coefficient respectively). 
In Figure \ref{fig:vog_plis_dp}, we see that some images, which we would expect to have large VoG scores (i.e. those that are less clear visually) posses the smallest PLIS values, meaning that the selection for images with particularly large/small VoG and PLIS could often be contradictory ($0.33$ and $0.34$ values of Pearson correlation coefficient respectively).
Furthermore, we see in Figure \ref{fig:plis_priv_nonpriv} that the images with  lowest PLIS scores are not consistent across DP and non-DP settings. 
This may suggest that while PLIS could be used to identify atypical images, it is unlikely to be consistent in identifying more difficult data samples when performing DP and non-DP training.

This raises an interesting issue: Images with high PLIS values are considered to have more revealing features and images with high VoG are considered to be difficult for the model to learn from. 
Intuitively, one would expect that more revealing features would correspond to rare, OOD samples \cite{mueller2022input}, making these features more uniquely identifying.
However, when comparing their PLIS and VoG values, we see that the model does not necessarily see them as more \say{challenging} to learn from.
One reason behind this might be that, similarly to \cite{zhang2020secret}, revealing features should correspond to more informative samples, as they contain more useful features for the model to learn from (and are considered to be \textit{easier}). 
And, as a result, high PLIS may indicate that the sample has rare features, which are easy for the model to learn from.
An in-depth investigation of \say{what makes a sample informative} and how it correlates to its difficulty is part of our ongoing work in this area. 

% \begin{figure}[h!]
% \centering
%       \includegraphics[width=0.45\linewidth]{figures/vog_plis_graph.png}
%     % \hspace{5mm}%
%       \includegraphics[width=0.45\linewidth]{figures/vog_plis_graph_vanilla.png}
% \caption{Comparison VoG and PLIS scores for a private ($\varepsilon=4$) and a non-private settings (ResNet-$18$, CIFAR-$10$).}
% \label{fig:graph_vog}
% \end{figure}

\begin{table*}[!ht]
\centering
\resizebox{0.78\textwidth}{!}{%
\begin{tabular}{@{}ccccc@{}}
\toprule
 & Loss values & Gradient norms & VoG & PLIS \\ \midrule
\begin{tabular}[c]{@{}c@{}}Test-time accuracy  $\varepsilon$=1 (\%)\end{tabular} & 56.6 & 55.1 & 52.4 & 49.9 \\
\begin{tabular}[c]{@{}c@{}}Test-time accuracy $\varepsilon$=4 (\%)\end{tabular} & 63.6 & 61.8 & 58.8 & 55.6 \\
\begin{tabular}[c]{@{}c@{}}Test-time accuracy $\varepsilon$=8 (\%)\end{tabular} & 66.7 & 64.8 & 61.7 & 58.0 \\ \bottomrule
\end{tabular} %
}
\caption{Removal of $25\%$ of all samples with highest metrics (ResNet-$18$, CIFAR-$10$).}
\label{tab:removal}
\end{table*}

\subsection{Importance sampling for model training}
In addition to incentivisation of the data owners, many of these metrics are often used for sample selection in decentralised model training, pruning the training set over time to remove redundancy and decrease training time.
% Instead of using all data points over training, the federation can remove a certain proportion of the dataset which was not beneficial for the modal or redundant after a certain number of rounds. 
We simulated such scenarios by training the model for $5$ to $10$ epochs depending on the complexity of the task and then removing a proportion (between $0.25$ and $0.35$) of the dataset based on the metrics of interest. 
The training then resumes (for $10$-$20$ epochs based on the complexity of the task).
We show our preliminary results in Table \ref{tab:removal} and reduce the overall training time by $50\%$.
Here we observe two distinct results when removing the most difficult samples (i.e those with the highest metrics). 
Firstly, for all metrics the train-time accuracy increases (and in most cases gets close to $100\%$ in non-private settings).
Secondly, for loss- and gradient-norm-based removal, the resulting test-time accuracy was on average much higher compared to PLIS and VoG-guided sample removal (across all $\varepsilon$ levels).
These results suggest that while \textit{some} information is lost by removing the samples identified as the more difficult ones through per-sample loss and gradient norm, these samples were, in fact, not the most informative ones.
On another hand, for PLIS and VoG-based removal, there was a significant drop in test-time accuracy across all privacy levels.
Therefore, we hypothesise that the difficulty captured through VoG and PLIS values is the so-called \say{useful difficulty}, where highest-ranking samples are beneficial for model generalisation.
Exploration of this phenomenon and its implications are part of our ongoing work in this domain.

% which suggest that removal of samples with high VoG or PLIS scores almost universally reduces the utility of the final model. 
% Moreover, we see that removing samples of low VoG or PLIS does not seem to reduce the utility of the model on the test set, but reduces the training time by a factor of $2$.
% suggesting that samples with largest per-sample losses and gradient norms do not necessarily represent that they are difficult 
% When comparing against the loss- or gradient-norm-based removal, we note that A) these metrics are not consistent between the private and the non-private settings and B) removal of samples with lowest loss or gradient norm values results in model overfitting (with train-time accuracy being close to $100\%$) and much worse (by up to $15\%$) test-time performance in DP models.
% These results suggest that PLIS and VoG metrics can be used in private decentralised sample selection effectively across all $\varepsilon$ levels. 

\subsection{Differentially private VoG and PLIS}
\label{sec:dp_vog}
As both the VoG and the PLIS values are sensitive, data-dependent quantities, these should not be published \say{in the clear}, as this can permit adversarial actors to identify potential victims (e.g. prioritising attacking high-valued participants).
One privacy-preserving alternative would be DP publication of these values.
For VoG values, we can perform differentially private variance queries ($\varepsilon, 0$) and for PLIS values we propose to use clipping and use a Laplace mechanism (also $\varepsilon, 0$).
These steps can then either be composed (heterogeneously) with the Gaussian Mechanism used for DP-SGD training, for instance using \cite{zhu2022optimal}, or have their own separate accountant. 

\section{Limitations and future work}
% The main difficulty of using these methods is the fact that not every difficult sample is equally useful for the model to learn from. As seen in Figure \ref{fig:pppd_epsilon}, high VoGs often correspond to samples on the tails of the data distribution, so learning from only these samples could be detrimental to the model (similar logic applying to the PLIS scores)
The main challenge of interpretation of these scores is the fact that not every difficult image is informative (and hence beneficial for the joint model). 
While our preliminary results suggest that VoG and PLIS scores tend to favor more useful samples, further work is needed to identify which features make a sample difficult and/or informative.
% : They often show samples, whose features are \say{unusual}, but not necessarily useful to the model. 

As we outlined in Section \ref{sec:dp_vog}, both the VoG and the PLIS values are sensitive, as they are derived directly from the private characteristics of the model. Therefore, these should only be shared in a privatised form, which could reduce their descriptiveness in settings with lower values of $\varepsilon$. 
This could lead to misalignment of the \textit{usefulness} of the data and the associated rewards. 
The investigation of which specific input features contribute towards higher \say{useful difficulty} of the sample is a promising area of future research. 
Investigation of the relationship between these metrics and model memorisation \cite{feldman2020does}, is part of our ongoing work.
Additionally, when studying the effects of federated retraining, extra care is required when calculating the privacy guarantees for the retrained model.
This is because both the sampling rates as well as the individuals in the dataset who benefit from the guarantees are directly affected.

As both the VoG and PLIS (and even the $L_2$-norm) calculations heavily rely on per-sample gradients, these methods are computationally limited by the speed of this calculation.
We see a noticeable overhead in performance, in some cases almost doubling the training time of the original model.
Therefore, gradient-based metrics can perform best in settings where dataset pruning results in a larger gain in performance than the computational overhead of these methods, as these metrics need to only be calculated once.
Our preliminary retraining results show that notable performance improvement can already be achieved using these metrics for data selection and we aim to continue this investigation in large-scale federated settings as part of the future work.

% Additionally, as these gradient-based methods are context agnostic (i.e. they can be used on any model-dataset combination), exploration of these techniques in other learning setting remains an open challenge. 
% While some of our results demonstrate that the trends in image selection remain identical across many models, it is not yet clear how a more complex task (or a dataset) interplays with a more parameterised model, which is part of our ongoing work in this area.

\section{Conclusion}
In this work, we show that it is possible to employ gradient-based metrics to identify samples of interest in private collaborative image analysis tasks.
We show that these can be effectively used to augment the FL imaging settings to permit more effective data selection.
These methods are economical computationally, effective across a range of different computer vision tasks on a variety of model architectures and perform well even in low-trust environments.
We note that VoG, while not originally designed with private training in mind, can be effectively used to identify such samples irrespective of the desired privacy level for a range of $\varepsilon$ in many learning settings. 
While PLIS was originally specifically developed to serve DP model training, we found that it could be less consistent when compared to VoG-based data selection depending on the value of $\varepsilon$.
A direct comparison between these two values shows that it is very challenging to precisely define the notion of a difficult and informative sample.
% The interplay between the feature complexity of an image and its usefulness in model training is a highly complex topic, which requires further investigation.
We hope that this work would encourage researchers to study the question of private-preserving data valuation further, encouraging further scientific collaboration and wider adoption of safe FL.  
% \printbibliography
%%
% \newpage

% \bibliography{iclr2024_conference}
% \bibliographystyle{iclr2024_conference}
\printbibliography

@article{glocker2023algorithmic,
  title={Algorithmic encoding of protected characteristics in chest X-ray disease detection models},
  author={Glocker, Ben and Jones, Charles and Bernhardt, M{\'e}lanie and Winzeck, Stefan},
  journal={Ebiomedicine},
  volume={89},
  year={2023},
  publisher={Elsevier}
}

@inproceedings{ahmad2020fairness,
  title={Fairness in machine learning for healthcare},
  author={Ahmad, Muhammad Aurangzeb and Patel, Arpit and Eckert, Carly and Kumar, Vikas and Teredesai, Ankur},
  booktitle={Proceedings of the 26th ACM SIGKDD International Conference on Knowledge Discovery \& Data Mining},
  pages={3529--3530},
  year={2020}
}

@article{mhasawade2021machine,
  title={Machine learning and algorithmic fairness in public and population health},
  author={Mhasawade, Vishwali and Zhao, Yuan and Chunara, Rumi},
  journal={Nature Machine Intelligence},
  volume={3},
  number={8},
  pages={659--666},
  year={2021},
  publisher={Nature Publishing Group UK London}
}

@article{geiping2020inverting,
  title={Inverting Gradients--How easy is it to break privacy in federated learning?},
  author={Geiping, Jonas and Bauermeister, Hartmut and Dr{\"o}ge, Hannah and Moeller, Michael},
  journal={arXiv preprint arXiv:2003.14053},
  year={2020}
}

@inproceedings{zhang2020secret,
  title={The secret revealer: Generative model-inversion attacks against deep neural networks},
  author={Zhang, Yuheng and Jia, Ruoxi and Pei, Hengzhi and Wang, Wenxiao and Li, Bo and Song, Dawn},
  booktitle={Proceedings of the IEEE/CVF Conference on Computer Vision and Pattern Recognition},
  pages={253--261},
  year={2020}
}

@article{konevcny2016federated,
  title={Federated learning: Strategies for improving communication efficiency},
  author={Kone{\v{c}}n{y}, Jakub and McMahan, H Brendan and Yu, Felix X and Richt{\'a}rik, Peter and Suresh, Ananda Theertha and Bacon, Dave},
  journal={arXiv preprint arXiv:1610.05492},
  year={2016}
}

@article{usynin2021adversarial,
  title={Adversarial interference and its mitigations in privacy-preserving collaborative machine learning},
  author={Usynin, Dmitrii and Ziller, Alexander and Makowski, Marcus and Braren, Rickmer and Rueckert, Daniel and Glocker, Ben and Kaissis, Georgios and Passerat-Palmbach, Jonathan},
  journal={Nature Machine Intelligence},
  volume={3},
  number={9},
  pages={749--758},
  year={2021},
  publisher={Nature Publishing Group}
}

@article{kaissis2021end,
  title={End-to-end privacy preserving deep learning on multi-institutional medical imaging},
  author={Kaissis, Georgios and Ziller, Alexander and Passerat-Palmbach, Jonathan and Ryffel, Th{\'e}o and Usynin, Dmitrii and Trask, Andrew and Lima, Ion{\'e}sio and Mancuso, Jason and Jungmann, Friederike and Steinborn, Marc-Matthias and others},
  journal={Nature Machine Intelligence},
  volume={3},
  number={6},
  pages={473--484},
  year={2021},
  publisher={Nature Publishing Group}
}

@inproceedings{abadi2016deep,
  title={Deep learning with differential privacy},
  author={Abadi, Martin and Chu, Andy and Goodfellow, Ian and McMahan, H Brendan and Mironov, Ilya and Talwar, Kunal and Zhang, Li},
  booktitle={Proceedings of the 2016 ACM SIGSAC conference on computer and communications security},
  pages={308--318},
  year={2016}
}

@article{usynin2022beyond,
  title={Beyond gradients: Exploiting adversarial priors in model inversion attacks},
  author={Usynin, Dmitrii and Rueckert, Daniel and Kaissis, Georgios},
  journal={arXiv preprint arXiv:2203.00481},
  year={2022}
}

@article{dpbook,
url = {http://dx.doi.org/10.1561/0400000042},
year = {2014},
volume = {9},
journal = {Foundations and Trends® in Theoretical Computer Science},
title = {The Algorithmic Foundations of Differential Privacy},
doi = {10.1561/0400000042},
issn = {1551-305X},
number = {3–4},
pages = {211-407},
author = {Cynthia Dwork and Aaron Roth}
}

@article{bagdasaryan2019differential,
  title={Differential privacy has disparate impact on model accuracy},
  author={Bagdasaryan, Eugene and Poursaeed, Omid and Shmatikov, Vitaly},
  journal={Advances in neural information processing systems},
  volume={32},
  year={2019}
}

@inproceedings{agarwal2022estimating,
  title={Estimating example difficulty using variance of gradients},
  author={Agarwal, Chirag and D'souza, Daniel and Hooker, Sara},
  booktitle={Proceedings of the IEEE/CVF Conference on Computer Vision and Pattern Recognition},
  pages={10368--10378},
  year={2022}
}

@article{mueller2022input,
  title={How Do Input Attributes Impact the Privacy Loss in Differential Privacy?},
  author={Mueller, Tamara T and Kolek, Stefan and Jungmann, Friederike and Ziller, Alexander and Usynin, Dmitrii and Knolle, Moritz and Rueckert, Daniel and Kaissis, Georgios},
  journal={arXiv preprint arXiv:2211.10173},
  year={2022}
}

@article{usynin2021distributed,
  title={Distributed Machine Learning and the Semblance of Trust},
  author={Usynin, Dmitrii and Ziller, Alexander and Rueckert, Daniel and Passerat-Palmbach, Jonathan and Kaissis, Georgios},
  journal={arXiv preprint arXiv:2112.11040},
  year={2021}
}

@article{de2022unlocking,
  title={Unlocking high-accuracy differentially private image classification through scale},
  author={De, Soham and Berrada, Leonard and Hayes, Jamie and Smith, Samuel L and Balle, Borja},
  journal={arXiv preprint arXiv:2204.13650},
  year={2022}
}

@article{xu2021gradient,
  title={Gradient driven rewards to guarantee fairness in collaborative machine learning},
  author={Xu, Xinyi and Lyu, Lingjuan and Ma, Xingjun and Miao, Chenglin and Foo, Chuan Sheng and Low, Bryan Kian Hsiang},
  journal={Advances in Neural Information Processing Systems},
  volume={34},
  pages={16104--16117},
  year={2021}
}

@inproceedings{jia2019towards,
  title={Towards efficient data valuation based on the shapley value},
  author={Jia, Ruoxi and Dao, David and Wang, Boxin and Hubis, Frances Ann and Hynes, Nick and G{"u}rel, Nezihe Merve and Li, Bo and Zhang, Ce and Song, Dawn and Spanos, Costas J},
  booktitle={The 22nd International Conference on Artificial Intelligence and Statistics},
  pages={1167--1176},
  year={2019},
  organization={PMLR}
}

@inproceedings{sim2020collaborative,
  title={Collaborative machine learning with incentive-aware model rewards},
  author={Sim, Rachael Hwee Ling and Zhang, Yehong and Chan, Mun Choon and Low, Bryan Kian Hsiang},
  booktitle={International Conference on Machine Learning},
  pages={8927--8936},
  year={2020},
  organization={PMLR}
}

@inproceedings{jia2021scalability,
  title={Scalability vs. utility: Do we have to sacrifice one for the other in data importance quantification?},
  author={Jia, Ruoxi and Wu, Fan and Sun, Xuehui and Xu, Jiacen and Dao, David and Kailkhura, Bhavya and Zhang, Ce and Li, Bo and Song, Dawn},
  booktitle={Proceedings of the IEEE/CVF Conference on Computer Vision and Pattern Recognition},
  year={2021}
}

@article{darlow2018cinic,
  title={Cinic-10 is not imagenet or cifar-10},
  author={Darlow, Luke N and Crowley, Elliot J and Antoniou, Antreas and Storkey, Amos J},
  journal={arXiv preprint arXiv:1810.03505},
  year={2018}
}

@article{yousefpour2021opacus,
  title={Opacus: User-friendly differential privacy library in PyTorch},
  author={Yousefpour, Ashkan and Shilov, Igor and Sablayrolles, Alexandre and Testuggine, Davide and Prasad, Karthik and Malek, Mani and Nguyen, John and Ghosh, Sayan and Bharadwaj, Akash and Zhao, Jessica and others},
  journal={arXiv preprint arXiv:2109.12298},
  year={2021}
}

@inproceedings{farrand2020neither,
  title={Neither private nor fair: Impact of data imbalance on utility and fairness in differential privacy},
  author={Farrand, Tom and Mireshghallah, Fatemehsadat and Singh, Sahib and Trask, Andrew},
  booktitle={Proceedings of the 2020 workshop on privacy-preserving machine learning in practice},
  pages={15--19},
  year={2020}
}

@inproceedings{zhu2022optimal,
  title={Optimal accounting of differential privacy via characteristic function},
  author={Zhu, Yuqing and Dong, Jinshuo and Wang, Yu-Xiang},
  booktitle={International Conference on Artificial Intelligence and Statistics},
  pages={4782--4817},
  year={2022},
  organization={PMLR}
}

@article{chen2020understanding,
  title={Understanding gradient clipping in private sgd: A geometric perspective},
  author={Chen, Xiangyi and Wu, Steven Z and Hong, Mingyi},
  journal={Advances in Neural Information Processing Systems},
  volume={33},
  pages={13773--13782},
  year={2020}
}

@inproceedings{feldman2020does,
  title={Does learning require memorization? a short tale about a long tail},
  author={Feldman, Vitaly},
  booktitle={Proceedings of the 52nd Annual ACM SIGACT Symposium on Theory of Computing},
  pages={954--959},
  year={2020}
}

@article{shobeiri2022shapley,
  title={Shapley Value is an Equitable Metric for Data Valuation},
  author={Shobeiri, Seyedamir and Aajami, Mojtaba},
  journal={Iraqi Journal for Electrical And Electronic Engineering},
  volume={18},
  number={2},
  year={2022},
  publisher={Basrah University}
}

@article{wang2004image,
  title={Image quality assessment: from error visibility to structural similarity},
  author={Wang, Zhou and Bovik, Alan C and Sheikh, Hamid R and Simoncelli, Eero P},
  journal={IEEE transactions on image processing},
  volume={13},
  number={4},
  pages={600--612},
  year={2004},
  publisher={IEEE}
}

@article{heusel2017gans,
  title={Gans trained by a two time-scale update rule converge to a local nash equilibrium},
  author={Heusel, Martin and Ramsauer, Hubert and Unterthiner, Thomas and Nessler, Bernhard and Hochreiter, Sepp},
  journal={Advances in neural information processing systems},
  volume={30},
  year={2017}
}

@article{kailath1967divergence,
  title={The divergence and Bhattacharyya distance measures in signal selection},
  author={Kailath, Thomas},
  journal={IEEE transactions on communication technology},
  volume={15},
  number={1},
  pages={52--60},
  year={1967},
  publisher={IEEE}
}

@inproceedings{xue2021toward,
  title={Toward understanding the influence of individual clients in federated learning},
  author={Xue, Yihao and Niu, Chaoyue and Zheng, Zhenzhe and Tang, Shaojie and Lyu, Chengfei and Wu, Fan and Chen, Guihai},
  booktitle={Proceedings of the AAAI Conference on Artificial Intelligence},
  volume={35},
  number={12},
  pages={10560--10567},
  year={2021}
}

@article{amiri2021convergence,
  title={Convergence of update aware device scheduling for federated learning at the wireless edge},
  author={Amiri, Mohammad Mohammadi and G{\"u}nd{\"u}z, Deniz and Kulkarni, Sanjeev R and Poor, H Vincent},
  journal={IEEE Transactions on Wireless Communications},
  volume={20},
  number={6},
  pages={3643--3658},
  year={2021},
  publisher={IEEE}
}

@article{baldock2021deep,
  title={Deep learning through the lens of example difficulty},
  author={Baldock, Robert and Maennel, Hartmut and Neyshabur, Behnam},
  journal={Advances in Neural Information Processing Systems},
  volume={34},
  pages={10876--10889},
  year={2021}
}

@inproceedings{lai2021oort,
  title={Oort: Efficient federated learning via guided participant selection},
  author={Lai, Fan and Zhu, Xiangfeng and Madhyastha, Harsha V and Chowdhury, Mosharaf},
  booktitle={15th $\{$USENIX$\}$ Symposium on Operating Systems Design and Implementation ($\{$OSDI$\}$ 21)},
  pages={19--35},
  year={2021}
}

@inproceedings{li2021sample,
  title={Sample-level data selection for federated learning},
  author={Li, Anran and Zhang, Lan and Tan, Juntao and Qin, Yaxuan and Wang, Junhao and Li, Xiang-Yang},
  booktitle={IEEE INFOCOM 2021-IEEE Conference on Computer Communications},
  pages={1--10},
  year={2021},
  organization={IEEE}
}

@inproceedings{ghorbani2019data,
  title={Data shapley: Equitable valuation of data for machine learning},
  author={Ghorbani, Amirata and Zou, James},
  booktitle={International conference on machine learning},
  pages={2242--2251},
  year={2019},
  organization={PMLR}
}
\appendix
\section{Appendix}

\subsection{Models and datasets}
\label{sec:settings}
Here we briefly outline the model architectures and datasets used in our study. We experiment with a number of ResNet-derived architectures, namely ResNet-$9,18$ as the \textit{smaller} shared models and ResNet-$101$, ResNeXt-$101$ and WideResNet-$50, 101$ as the \textit{larger} models. We use the ResNet model family, as it was previously shown to be more robust to clipping (required for DP-SGD) and has obtained very adequate accuracy in the past on a variety of computer vision tasks in both the private and the non-private training settings \cite{de2022unlocking}. 

In this work we employ the commonly used CIFAR-$10$ dataset as well as two more complex tasks, namely the CINIC-$10$ \cite{darlow2018cinic} and the paediatric pneumonia prediction (PPPD) (adapted from \cite{kaissis2021end}) datasets. Both CIFAR-$10$ ($50\,000$ of $32x32$ training images) and CINIC-$10$ ($90\,000$ of $32x32$ training images) are classification tasks with $10$ classes. PPPD ($5\,400$ of $224x224$ training images) is a classification task with $3$ classes: no pathology, bacterial pneumonia and viral pneumonia. 
\subsection{Experimental settings}
We perform model training using PyTorch $1.13$, DP training using \texttt{opacus} $1.3.0$ \cite{yousefpour2021opacus} on two Linux $22.04$ machines using NVIDIA RTX A$6\, 000$ and NVIDIA Quadro RTX $5\,000$ GPUs respectively. To obtain fast per-sample gradients we employ the \texttt{functorch} $1.13.1$ library. 
For differentially private training we use three distinct privacy settings, with $\varepsilon$ values of $1$, $4$ and $8$. Our $\delta$ values are dataset-defined with $\delta_{\text{CIFAR}}=1e^{-5}$, $\delta_{\text{CINIC}}=1.1e^{-6}$ and $\delta_{\text{PPPD}}=1e^{-4}$

\end{document}